\documentclass[12pt, letterpaper]{article}

\usepackage[margin=1in]{geometry}
\usepackage{times}
\usepackage{graphicx}
\usepackage{float}
\usepackage{amsmath, amssymb}
\usepackage{booktabs}
\usepackage[numbers, sort&compress]{natbib}
\usepackage[colorlinks=true, linkcolor=blue, citecolor=blue, urlcolor=blue]{hyperref}
\usepackage{subcaption}
\usepackage{caption}
\title{Analyzing Visual Aircraft Representations\\with Sparse Autoencoders}

\author{
	\normalsize Deepshik Sharma\\
	{\normalsize Department of Computer Science \& Engineering}\\ 
	{\normalsize Jain University, Bangalore, India}\\
	{\small \texttt{21btrca058@jainuniversity.ac.in}}\\ 
}

\date{}  

\begin{document}
	
	\maketitle

	\begingroup
	\renewcommand{\thefootnote}{}
	\footnotetext{Code is available at \url{https://github.com/deepshiksharma/sae-aircraft}\par\smallskip}
	\footnotetext{An earlier version of this work was submitted to Jain University in May 2025, in partial fulfillment of degree requirements for the Bachelor of Technology program.}
	\addtocounter{footnote}{-1}
	\endgroup
	
	\begin{abstract}
Vision models can achieve strong performance on classification tasks, but the internal representations supporting their predictions are often difficult to interpret. This work investigates whether sparse autoencoders can decompose intermediate representations of a vision model into interpretable features. We train a ConvNeXt classifier on the FGVC-Aircraft dataset, extract spatial activations from its final feature stage, and train a sparse autoencoder on these activations. The learned sparse features are analyzed using top-activating image patches, activation strength, and class selectivity. Qualitative visual inspection reveals that several features correspond to recognizable aircraft structures and visual patterns. We evaluate a subset of selected features using input-space and feature-space ablations, measuring how blurring image patches and suppressing sparse features affect class logits, classification margins, and prediction confidence. The results suggest that sparse autoencoders can reveal partially interpretable, class-relevant visual features associated with aircraft recognition, while also exposing limitations such as polysemanticity and coarse spatial localization.
\end{abstract}

	\vspace{1em}  
	
	\section{Introduction}
The strong predictive performance of deep vision models alone does not explain which internal representations support a model's decisions. In an aircraft classification task, visually similar aircraft types may differ only through subtle structural cues such as airframe geometry and configuration \cite{maji2013fgvcaircraft}. A model may rely on these cues in ways that are difficult to understand from the final prediction alone.\\

A common approach to interpreting image classifiers is to use attribution methods such as saliency maps, CAM, Grad-CAM, or related variants \cite{simonyan2013saliency, zhou2016cam, selvaraju2017gradcam, chattopadhyay2018gradcampp, wang2020scorecam}. These methods highlight image regions that are influential for a particular class prediction by estimating how strongly different spatial regions contribute to the model's output score for a target class. Such methods are useful for visualizing where the evidence for a class is concentrated, but they do not directly explain how that evidence is represented inside the model. For example, a heatmap over the fuselage or wing region does not indicate whether the model has learned distinct reusable features corresponding to aircraft geometry, configuration, markings, or other correlated visual context.

Concept-based interpretability methods address this limitation more directly by analyzing internal activations rather than only input-space attribution. TCAV (Testing with Concept Activation Vectors) relates class predictions to user-defined concept directions \cite{kim2018tcav}, Network Dissection assigns semantic labels to hidden units \cite{bau2017networkdissection}, and concept bottleneck models explicitly predict human-defined concepts before classification \cite{koh2020conceptbottleneck}. However, these approaches often depend on predefined concepts and concept annotations. In many cases, this can be restrictive because relevant cues may be local, entangled, dataset-specific, or difficult to specify exhaustively in advance.\\

Sparse autoencoders (SAEs) provide an alternative way to analyze internal representations by learning a sparse decomposition of activation vectors without requiring predefined concept labels \cite{olshausen1996sparsecoding, elhage2022superposition, bricken2023monosemanticity, templeton2024scalingmonosemanticity, stevens2025interpretabletestablevisionfeatures}. Rather than assigning importance to image pixels for a single prediction, a sparse autoencoder learns latent features that can recur across images and spatial locations, enabling analysis of recurring visual patterns in the model's intermediate representations.

This work investigates whether SAEs can reveal interpretable, class-relevant features in the intermediate representations of an aircraft image classification model. We train ConvNeXt-Tiny \cite{liu2022convnext} on the FGVC-Aircraft dataset \cite{maji2013fgvcaircraft}, extract spatial activations from the model's final feature stage, and train a sparse autoencoder on the extracted patch-level activation vectors. We then analyze the learned sparse features using top-activating image patches, activation strength, and class selectivity. We find that several sparse features correspond in part to visually meaningful aircraft structures, including landing gear, flap track fairings, fuselage windows, cockpit region, engine \& wing adjacent regions, and biplane struts. Finally, we evaluate a subset of eight selected features using input-space ablations based on localized image-region perturbations, and feature-space ablations based on latent feature suppression, testing whether the visual regions and latent features identified by the SAE influence class logits, classification margins, and prediction confidence. Overall, the results suggest that SAEs can reveal partially interpretable aircraft-related features in intermediate representations. At the same time, these results also suggest that these features are not perfectly semantic: some are polysemantic, some correspond to part-context combinations, and prediction-relevant information is often distributed across multiple spatial regions and sparse features.

	\section{Related Work}

\subsection{Attribution and Localization Methods}

Attribution and localization methods are commonly used to explain image classifier predictions by identifying input regions that influence a target class score. Early saliency methods used gradients of class scores with respect to input pixels to highlight regions relevant to a prediction \cite{simonyan2013saliency}. Integrated Gradients later proposed an axiomatic gradient-based attribution method that accumulates gradients along a path from a baseline input to the original input \cite{sundararajan2017integratedgradients}. CAM showed that convolutional neural networks with global average pooling can produce class-discriminative localization maps from final convolutional feature maps \cite{zhou2016cam}. Grad-CAM generalized this approach by using gradients flowing into convolutional feature maps to generate coarse class-specific heatmaps without requiring architectural modifications \cite{selvaraju2017gradcam}. Subsequent methods such as Grad-CAM++ and Score-CAM modify the weighting strategy or reduce reliance on gradients \cite{chattopadhyay2018gradcampp, wang2020scorecam}.

These methods are useful for visualizing where class-discriminative evidence is concentrated in an image. However, they primarily provide input-space or feature-map localization for a target prediction. They do not directly decompose internal representations into reusable latent features that can be analyzed across images, spatial positions, and classes.

\subsection{Concept-Based Interpretability}

Concept-based interpretability methods study model behavior in terms of human-interpretable concepts. TCAV (Testing with Concept Activation Vectors) measures the sensitivity of class predictions to user-defined concept directions in activation space \cite{kim2018tcav}. Network Dissection quantifies the interpretability of hidden units by measuring their alignment with semantic concepts such as objects, parts, textures, and materials \cite{bau2017networkdissection}. Concept bottleneck models make concepts explicit by predicting human-defined concept variables before producing the final class prediction, allowing interventions at the concept level \cite{koh2020conceptbottleneck}.

These methods provide a useful bridge between raw neural activations and human-interpretable descriptions. However, many concept-based methods require predefined concepts along with concept annotations or labeled examples. This can be restrictive when relevant visual cues are local, entangled, dataset-specific, or not known exhaustively in advance.

\subsection{Mechanistic Interpretability and Sparse Autoencoders}

Mechanistic interpretability aims to understand neural networks by identifying internal components that implement meaningful parts of a model's computation, rather than only describing input-output behavior \cite{olah2020circuits, elhage2022superposition}. In this setting, a useful explanation should not only assign importance to an input region, but should also relate model behavior to internal activations that can be inspected, interpreted, and intervened on.\\

Sparse coding and sparse dictionary learning provide one way to decompose representations into simpler components. Sparse coding represents a dense input vector as a combination of a small number of learned basis elements, called dictionary atoms \cite{olshausen1996sparsecoding}. This is useful for neural network interpretability because individual neurons may be polysemantic: the same neuron can respond to multiple unrelated patterns. Work on superposition argues that neural networks may represent more features than they have individual neurons by encoding features as directions in activation space rather than as isolated units \cite{elhage2022superposition}.\\

Sparse autoencoders adapt the sparse decomposition idea using an encoder-decoder architecture: the encoder maps dense activation vectors into an overcomplete sparse latent space, while the decoder reconstructs the original activation from those sparse latents. In mechanistic interpretability, SAEs have been used to decompose dense neural activations into sparse latent features that can be more interpretable than individual neurons \cite{bricken2023monosemanticity, templeton2024scalingmonosemanticity}.

Recent work has extended SAE-based analysis to vision models. Stevens et al. apply sparse autoencoders to vision representations and emphasize that feature interpretation should be supported not only by visualizing top-activating examples, but also by controlled interventions on learned features \cite{stevens2025interpretabletestablevisionfeatures}. This is particularly relevant for vision, where visually coherent top-activating examples may not by themselves establish whether a feature has causal influence on downstream model behavior. This line of work motivates our use of sparse autoencoders as a tool for analyzing internal representations of an aircraft image classification model.

	\section{Method}

\subsection{Dataset}
We use the FGVC-Aircraft dataset \cite{maji2013fgvcaircraft}, which contains 10,000 aircraft images organized in a hierarchy of manufacturer, family, and variant labels. We use family-level labels, which provide an intermediate level of granularity for aircraft type classification. Images are cropped using the provided bounding box coordinates. We use a train-validation-test split of 80\% / 10\% / 10\%, stratified by aircraft family.
\begin{figure}[H]
	\centering
	\includegraphics[width=\textwidth]{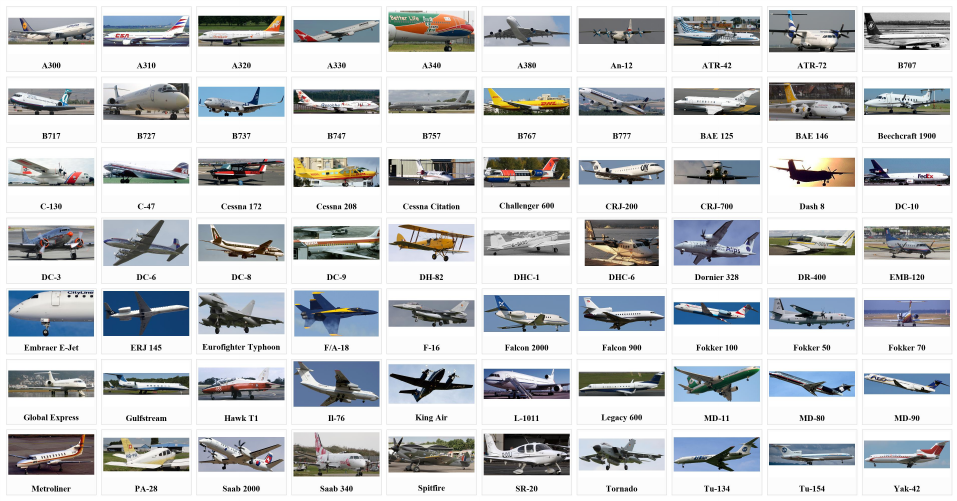}
	\caption{Sample images from each class in the 70 family-level classes in FGVC-Aircraft, illustrating variation in aircraft configuration, viewpoint, airline livery, and background context. Some classes are visually similar and differ only through subtle structural characteristics, making classification challenging.}
	\label{fig:fgvc-aircraft}
\end{figure}

\subsection{Classifier Training}
We chose ConvNeXt-Tiny \cite{liu2022convnext} as the classifier model, initialized with ImageNet-pretrained weights \cite{deng2009imagenet} and fine-tuned on the FGVC-Aircraft dataset. We train the model using cross-entropy loss, AdamW optimization with an initial learning rate of $5\times10^{-5}$, cosine learning rate decay to a minimum learning rate of $10^{-6}$, and a batch size of $32$. We select the checkpoint with the highest validation accuracy, which occurs at epoch 12. The selected model achieves an accuracy of $92.2\%$ and a macro F1 score of $92.0\%$ on the test set. These results indicate that the model learns discriminative aircraft representations with balanced performance across aircraft classes, making it a competent source model for representation analysis. This model is frozen and used as the representation source for all SAE-based analysis.

\subsection{Activation Extraction}
We extract activations from the output of the final feature stage of ConvNeXt-Tiny, immediately before global average pooling. A forward hook is registered on the corresponding module, and the frozen classifier is run in evaluation mode over each dataset split. This layer produces a spatial activation tensor of shape $(768 \times 7 \times 7)$. The $7 \times 7$ dimensions correspond to spatial positions in the feature map, while the $768$ channels represent the learned feature vector for each spatial location.\\

\begin{figure}[H]
    \centering
    \includegraphics[width=\textwidth]{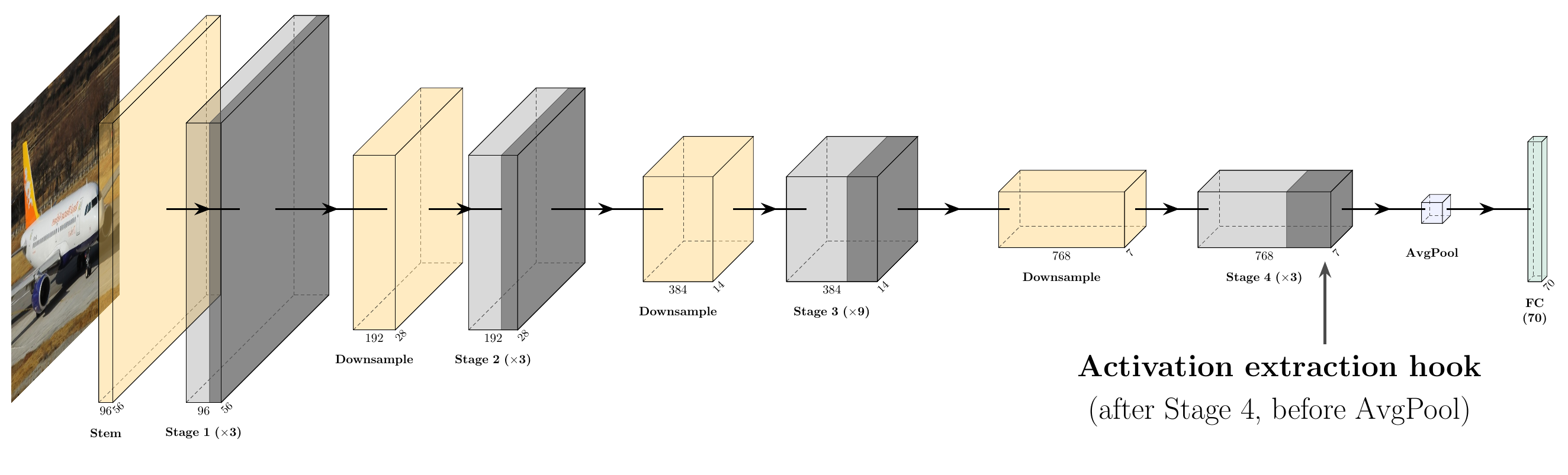}
	\caption{Placement of the forward hook used for activation extraction on ConvNeXt-Tiny.\\ \textit{\footnotesize This diagram was created using PlotNeuralNet~\cite{iqbal2018plotneuralnet}}}
    \label{fig:model-hooked}
\end{figure}

For each image, every spatial location is treated as a separate patch-level activation vector. We permute each activation tensor from $(C \times H \times W)$ to $(H \times W \times C)$ and reshape it into $(HW) \times C$. Since $H=W=7$ and $C=768$, each image produces $49$ vectors in $\mathbb{R}^{768}$. We apply this procedure separately to the training, validation, and test splits and save the resulting activation vectors for SAE training, model selection, and evaluation.

\subsection{Sparse Autoencoder Training}
The sparse autoencoder is trained on activations produced by images in the training split, while validation and test split activations are used for model validation and evaluation.\\

Let $x \in \mathbb{R}^{768}$ denote an image patch activation vector. The sparse autoencoder maps $x$ to an overcomplete sparse representation $z \in \mathbb{R}^{8192}$ using a ReLU encoder:
\[
z = \mathrm{ReLU}(W_{\mathrm{enc}}x + b_{\mathrm{enc}})
\]

The decoder reconstructs the original activation vector $x$ from $z$:
\[
\hat{x} = W_{\mathrm{dec}}z + b_{\mathrm{dec}}
\]

The sparse autoencoder is trained with a reconstruction loss and a sparsity penalty on the sparse feature activations. The training objective is
\[
\mathcal{L}
=
\mathcal{L}_{\mathrm{recon}}
+
\lambda \mathcal{L}_{\mathrm{sparse}}
\]

The reconstruction term, \(\mathcal{L}_{\mathrm{recon}} = \mathrm{MSE}(\hat{x}, x)\), penalizes differences between the reconstructed activation \(\hat{x}\) and the original activation \(x\).\\

The sparsity term, \(\mathcal{L}_{\mathrm{sparse}} = \mathrm{mean}(|z|)\), penalizes the average magnitude of the sparse feature activations. This encourages the autoencoder to represent each input using only a small number of active latent features.\\

In our experiment, we use \(\lambda = 10^{-2}\), so the training objective becomes
\[
\mathcal{L}
=
\mathrm{MSE}(\hat{x}, x)
+
10^{-2}\,\mathrm{mean}(|z|)
\]

Before training, the activations are centered using the mean of the training activations. The same mean is also used during validation, evaluation, feature discovery, and ablations.

Centering removes the mean activation component, so the SAE learns features that capture variation around the average activation pattern rather than a shared offset across examples.\\

The SAE is trained for 100 epochs, achieving a low reconstruction error and an explained variance score of 0.9978 on the test split activations. This indicates that the sparse representation preserves most of the information from the original activations.

We use the explained variance score as the primary measure of reconstruction quality, because the activations were mean-centered but not variance-normalized, so reconstruction error is a scale-dependent quantity.

\subsection{Feature Discovery}
After the SAE has been trained, we pass each patch-level activation vector from the test set through the SAE encoder to obtain sparse latent activations. For each SAE feature, we rank all test set spatial patches by that feature's activation value and retain the top $k=50$ activating patches. To prioritize features for qualitative inspection, we rank SAE features by the mean activation over their top-$50$ patches and select the top-$100$ SAE features under this criterion.

For each SAE feature in the top-$100$, we save its top-$50$ activating image patches along with their corresponding source images, allowing the feature to be inspected both in local visual context and in the context of the complete aircraft image. These saved examples form the basis for assigning qualitative interpretations to SAE features.

\subsubsection{Selecting SAE Features for Ablation Experiments}
For the ablation experiments, we select eight SAE features that are visually interpretable and cover a range of aircraft-related cues. The selected features include examples associated with the landing gear, flap track fairings, fuselage windows, cockpit region, engine \& wing adjacent regions, and biplane struts. We deliberately choose features with varied visual characteristics rather than selecting features solely by activation strength or class specificity. This allows the ablation experiments to evaluate whether several qualitatively different SAE features correspond to prediction-relevant evidence. For each selected feature, the ablation experiments are run on only the top-$50$ activating test set patches saved during feature discovery. This focuses the evaluation on examples where the selected feature is strongly active.

\subsection{Ablation Experiments}
We evaluate the selected SAE features using two ablation strategies. Input-space ablation tests whether the image regions associated with high SAE feature activation are important to the classifier. Feature-space ablation tests whether suppressing the corresponding SAE latent feature affects classifier behavior. Together, these two ablation strategies distinguish between visual relevance in the input image and the direct effects of modifying the SAE-decomposed internal representation of the vision model.

\subsubsection{Input-Space Ablation}
We construct four perturbed versions of each image, where the following patches are blurred, respectively: (a) the target patch, (b) an expanded target region, (c) a randomly selected non-target patch, (d) an expanded random region. The expanded region is defined as the $3 \times 3$ neighborhood centered on the selected patch, clipped at the image boundary.

We use Gaussian blur rather than replacing the image region with a constant color, since blurring suppresses local visual detail while reducing the introduction of sharp artificial artifacts. The random-region blurred images serve as controls for the effect of perturbing an arbitrary region of the same image. For each original and blurred image pair, we record the selected SAE feature activation at the target spatial location, true-class logit, true-class probability, classification margin, confidence assigned to the original top-1 class, and whether the predicted class changes. We compare target-region perturbations against their random-region controls to test whether SAE-selected regions have stronger effects on the classifier than unrelated image regions.

\subsubsection{Feature-Space Ablation}
We run each image through the ConvNeXt feature extractor to obtain the activation tensor from its final feature stage. We then encode the relevant activation vector using the SAE, set the selected latent feature activation to zero, decode the modified sparse representation back into ConvNeXt activation space, and continue the forward pass through the global average pooling and classifier layers.

We evaluate two types of feature-space interventions. Local intervention suppresses the selected SAE feature only at its single top-most activating spatial location. Global intervention suppresses the same SAE feature across all $49$ spatial locations in the $7 \times 7$ feature map. For both interventions, we compare target-feature suppression against a random-feature suppression control. We also include SAE reconstruction baselines in which activations are encoded and decoded without suppressing any latent feature. These baselines separate the effect of suppressing a specific SAE feature from the distortion introduced by SAE reconstruction.

For each intervention, we measure changes in true-class logit, true-class probability, classification margin, confidence assigned to the original top-1 class, and whether the predicted class changes. These metrics allow us to assess whether the selected SAE feature and its corresponding image region are associated with evidence used by the classifier.

\subsection{Evaluation Metrics}
We evaluate ablations by measuring how each intervention changes the classifier's output relative to the original, unablated image. The main metrics of interest are the true-class logit, the classification margin, the true-class probability, the confidence assigned to the original top-1 prediction, and the prediction flip rate.\\

The true-class logit is the classifier's raw output score for the ground-truth class. A decrease in this logit indicates reduced classifier evidence for the correct class.\\

The classification margin is defined as
\[
m(x) = f_y(x) - \max_{c \neq y} f_c(x),
\]
where \(f_y(x)\) is the true-class logit and \(\max_{c \neq y} f_c(x)\) is the highest competing class logit. A decrease in classification margin indicates that the correct class becomes less separated from the strongest alternative class.\\

The true-class probability is the softmax probability assigned to the ground-truth class. A decrease in this probability indicates that the classifier assigns less confidence to the correct class after ablation.\\

The original top-1 confidence is the probability assigned to the class predicted for the original, unablated image. A decrease in this confidence indicates that the ablation weakens the model's original prediction, even if the predicted class does not change.\\

The prediction flip rate is the fraction of examples for which the top-1 predicted class changes after ablation. A higher flip rate indicates that the intervention more often changes the classifier's final decision.\\

For each intervention, we compute drops relative to the original image. For a metric \(s\), the drop under intervention \(a\) is
\[
\Delta_a s = s(x) - s(x_a),
\]
where \(x\) is the original image or activation representation and \(x_a\) is the corresponding ablated version. Larger positive drops indicate a stronger reduction in the measured quantity.

	\section{Results}

\subsection{Qualitative SAE Feature Interpretations}

\begin{figure}[H]
	\centering
	\includegraphics[width=\textwidth]{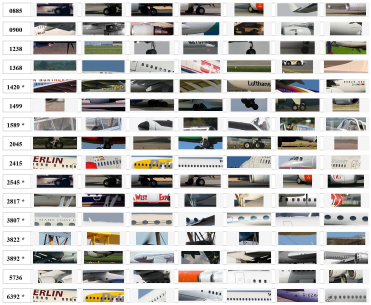}
	\caption{Sample images of the top-activating patches from a subset of 16 SAE features, including the 8 features used for ablation experiments (feature IDs are marked with * to indicate that the feature was selected for ablation).}
	\label{fig:sparse-features}
\end{figure}

Qualitative inspection of the top-activating patches shows that several SAE features correspond to recognizable aircraft structures and recurring visual patterns. Figure~\ref{fig:sparse-features} shows representative top-activating patches for a subset of 16 SAE features, including the eight features selected for ablation experiments. The selected features cover a range of aircraft-related structures, including landing gear, flap track fairings, fuselage windows, cockpit region, engine \& wing adjacent regions, and biplane struts, belonging to a wide range of aircraft types.\\

\begin{table}[H]
\centering
\small

\begin{subtable}{\textwidth}
\centering
\begin{tabular}{lp{0.72\textwidth}}
\toprule
Feature ID & Qualitative interpretation of feature \\
\midrule
$1420$ & Dominant pattern of flap track fairings; Secondary activations on various regions of fuselage \\
$1589$ & Dominant pattern of cockpit region of old warplanes; Occasional activations on the nose section \\
$2545$ & Dominant pattern of rear landing gear; Secondary activations on the engine region; exclusively B737 \\
$2817$ & Rear landing gear strut, and various fuselage \& engine-adjacent regions; exclusively Dash 8 \\
$3807$ & Dominant pattern of fuselage windows on business jets \\
$3822$ & Dominant pattern of struts in between the upper and lower wings of biplanes \\
$3892$ & Dominant pattern of wing and engine-adjacent regions; exclusively B747 \\
$6392$ & Dominant pattern of fuselage windows on commercial airliners \\
\bottomrule
\end{tabular}
\caption{Qualitative interpretations of selected SAE features.}
\label{tab:features-qualitative-interp}
\end{subtable}

\vspace{1.10em}

\begin{subtable}{\textwidth}
\centering
\begin{tabular}{rrrrrr}
\toprule
Feature ID & Mean Top-50 Actv. & Max Actv. & Top class & Top-class Frac. & Class Entropy \\
\midrule
$1420$ & 0.631 & 0.823 & A330 & 0.36 & 0.461 \\
$1589$ & 0.670 & 0.776 & DHC-1 & 0.78 & 0.171 \\
$2545$ & 0.729 & 0.746 & Boeing 737 & 1.00 & 0.000 \\
$2817$ & 0.718 & 0.657 & Dash 8 & 1.00 & 0.000 \\
$3807$ & 0.693 & 0.822 & Gulfstream & 0.90 & 0.092 \\
$3822$ & 0.688 & 0.791 & DH-82 & 0.98 & 0.023 \\
$3892$ & 0.717 & 0.655 & Boeing 747 & 1.00 & 0.000 \\
$6392$ & 1.000 & 1.000 & Boeing 737 & 0.78 & 0.207 \\
\bottomrule
\end{tabular}
\caption{Activation and class-selectivity statistics of selected SAE features. Mean top-50 activation, max activation, and class entropy are each scaled independently to the range $[0,1]$. A lower value for class entropy indicates greater class selectivity. Top class is the most frequently occurring class among the top-50 activating patches. Top-class fraction is the fraction of top-50 activating patches belonging to the most frequent class.}

\label{tab:features-showcase-stats}
\end{subtable}

\caption{Qualitative interpretations and summary statistics of SAE features selected for ablations.}
\label{tab:selected-features-summary}
\end{table}

Table~\ref{tab:selected-features-summary} summarizes the eight features selected for ablation experiments. A few features show strong class concentration among their top-activating patches. For example, feature IDs $2545$, $2817$, and $3892$ activate exclusively on the Boeing 737, Dash 8, and Boeing 747 examples respectively in their top-50 patches. Similarly, feature IDs $3807$ and $3822$ are heavily concentrated on the Gulfstream and DH-82 examples respectively. Other features are less class-specific but still visually interpretable. For example, feature ID $1420$, frequently activates on flap track fairings but also appears on other fuselage regions, while feature ID $6392$ activates on passenger-window patterns across commercial airliners.\\

A consistent pattern in these examples is that the selected features are not perfectly monosemantic. Although each feature has a dominant visual theme, none activates exclusively on a single clean aircraft part across all top examples. For example, the features associated with landing gear also sometimes activate on nearby engine or lower-fuselage regions, while features associated with fuselage windows may also include surrounding aircraft body structures. Thus, the qualitative labels in Table~\ref{tab:selected-features-summary}\subref{tab:features-qualitative-interp} should be read as approximate descriptions of dominant activation patterns rather than exact semantic labels.

This polysemanticity is not surprising, due to the SAE being trained on late ConvNeXt activations rather than on pixel-level part annotations. Its features therefore reflect directions in the model's internal representation space, not a manually defined aircraft-part taxonomy. The learned sparse features appear partially interpretable, but they are not cleanly aligned with isolated human-defined aircraft parts.

\subsection{Input-Space Ablation Results}

\begin{figure}[H]
    \centering
    \includegraphics[width=\textwidth]{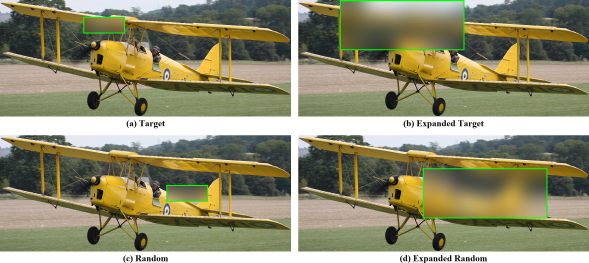}
	\caption{Perturbed images of a sample image selected for input-space ablation.}
    \label{fig:input-ablations}
\end{figure}

\begin{table}[H]
\centering
\small
\begin{tabular}{lrrrr}
\toprule
Metric & Target & Random & Expanded Target & Expanded Random \\
\midrule
SAE feature activation drop & 0.310 & 0.033 & 1.578 & 0.286 \\
True-class logit drop & 0.478 & 0.080 & 4.503 & 1.074 \\
True-class probability drop & 0.020 & 0.005 & 0.321 & 0.060 \\
Classification margin drop & 0.623 & 0.110 & 4.909 & 1.106 \\
Original top-1 confidence drop & 0.018 & 0.004 & 0.322 & 0.059 \\
Prediction flip rate & 0.015 & 0.005 & 0.220 & 0.045 \\
\bottomrule
\end{tabular}
\caption{A summary of input-space ablation results averaged over the eight selected SAE features. Target interventions blur the SAE-selected region, while random interventions blur a randomly selected non-target region from the same image. Larger drops indicate stronger reductions in classifier evidence.}
\label{tab:input_ablation_summary}
\end{table}

The input-space ablation results show that blurring SAE-selected regions has a larger effect on classifier behavior than blurring randomly selected regions from the same images. The results in Table~\ref{tab:input_ablation_summary} indicate that target patch blurring produces larger drops than random patch blurring across all reported metrics.

The effect becomes substantially stronger in the expanded-region setting, where the true-class logit drop increases from $1.074$ for expanded random regions to $4.503$ for expanded target regions, and the classification margin drop increases from $1.106$ to $4.909$.\\

The feature-wise results in Tables~\ref{tab:input_ablation_single_patch_by_feature}
and~\ref{tab:input_ablation_expanded_by_feature}
show that the observed pattern is broadly consistent across the selected SAE features, with the largest effects occurring in the expanded-region setting. For both single-patch and expanded-region blurring, target interventions usually produce larger reductions in SAE feature activation, true-class logits, probabilities, margins, and original top-1 confidence than the corresponding random interventions. The expanded-region setting produces the clearest effects, suggesting that the visual evidence associated with an SAE feature is often distributed over the selected patch and its surrounding spatial context rather than being isolated to a single feature-map location.

\begin{table}[H]
\centering
\small

\begin{subtable}{\textwidth}
\centering
\small
\resizebox{\textwidth}{!}{
\begin{tabular}{rrrrrrrrrrr}
\toprule
Feature ID &
\multicolumn{2}{c}{SAE feature actv. drop} &
\multicolumn{2}{c}{True-class logit drop} &
\multicolumn{2}{c}{True-class prob. drop} &
\multicolumn{2}{c}{Classification margin drop} &
\multicolumn{2}{c}{Original top-1 conf. drop} \\
\cmidrule(lr){2-3}
\cmidrule(lr){4-5}
\cmidrule(lr){6-7}
\cmidrule(lr){8-9}
\cmidrule(lr){10-11}
& Target & Random & Target & Random & Target & Random & Target & Random & Target & Random \\
\midrule
$1420$ & 0.382 & 0.018 & 0.622 & 0.058 & 0.051 & 0.001 & 0.669 & 0.097 & 0.040 & 0.000 \\
$1589$ & 0.292 & 0.007 & 0.514 & 0.077 & 0.034 & 0.007 & 0.683 & 0.129 & 0.034 & 0.007 \\
$2545$ & 0.832 & 0.025 & 0.784 & 0.092 & 0.002 & 0.003 & 1.395 & 0.154 & 0.002 & 0.003 \\
$2817$ & 0.204 & 0.042 & 0.328 & 0.070 & 0.001 & 0.000 & 0.248 & 0.047 & 0.001 & 0.000 \\
$3807$ & 0.132 & 0.070 & 0.269 & 0.129 & 0.005 & 0.020 & 0.428 & 0.189 & 0.005 & 0.020 \\
$3822$ & 0.183 & 0.032 & 0.250 & 0.086 & 0.011 & 0.001 & 0.467 & 0.133 & 0.011 & 0.001 \\
$3892$ & 0.166 & 0.012 & 0.590 & 0.040 & 0.020 & 0.000 & 0.712 & -0.007 & 0.020 & 0.000 \\
$6392$ & 0.290 & 0.057 & 0.467 & 0.089 & 0.035 & 0.004 & 0.378 & 0.139 & 0.032 & 0.004 \\
\bottomrule
\end{tabular}
}
\caption{Single-patch blurring. Target interventions blur the SAE-selected patch, while random interventions blur a randomly selected non-target patch from the same image.}
\label{tab:input_ablation_single_patch_by_feature}
\end{subtable}

\vspace{1.10em}

\begin{subtable}{\textwidth}
\centering
\small
\resizebox{\textwidth}{!}{
\begin{tabular}{rrrrrrrrrrr}
\toprule
Feature ID &
\multicolumn{2}{c}{SAE feature actv. drop} &
\multicolumn{2}{c}{True-class logit drop} &
\multicolumn{2}{c}{True-class prob. drop} &
\multicolumn{2}{c}{Classification margin drop} &
\multicolumn{2}{c}{Original top-1 conf. drop} \\
\cmidrule(lr){2-3}
\cmidrule(lr){4-5}
\cmidrule(lr){6-7}
\cmidrule(lr){8-9}
\cmidrule(lr){10-11}
& Target & Random & Target & Random & Target & Random & Target & Random & Target & Random \\
\midrule
$1420$ & 1.461 & 0.248 & 5.846 & 1.111 & 0.588 & 0.091 & 6.294 & 1.355 & 0.590 & 0.081 \\
$1589$ & 1.531 & 0.256 & 4.765 & 1.205 & 0.438 & 0.090 & 4.795 & 1.167 & 0.438 & 0.090 \\
$2545$ & 1.807 & 0.142 & 4.303 & 0.827 & 0.139 & 0.035 & 5.693 & 1.016 & 0.139 & 0.035 \\
$2817$ & 1.385 & 0.305 & 5.644 & 1.179 & 0.426 & 0.045 & 5.401 & 1.036 & 0.426 & 0.045 \\
$3807$ & 1.625 & 0.372 & 4.221 & 1.240 & 0.321 & 0.080 & 5.088 & 1.297 & 0.321 & 0.080 \\
$3822$ & 1.220 & 0.226 & 2.691 & 0.883 & 0.162 & 0.030 & 3.371 & 0.907 & 0.162 & 0.030 \\
$3892$ & 1.308 & 0.253 & 4.629 & 1.008 & 0.300 & 0.059 & 4.622 & 0.867 & 0.300 & 0.059 \\
$6392$ & 2.288 & 0.486 & 3.925 & 1.138 & 0.196 & 0.053 & 4.009 & 1.203 & 0.198 & 0.053 \\
\bottomrule
\end{tabular}
}
\caption{Expanded-region blurring. Target interventions blur the $3 \times 3$ neighborhood around the SAE-selected patch, while random interventions blur the $3 \times 3$ neighborhood around a randomly selected non-target patch from the same image.}
\label{tab:input_ablation_expanded_by_feature}
\end{subtable}

\caption{Per-feature input-space ablation results.}
\label{tab:input_ablation_by_feature}
\end{table}

Prediction flip rates shown in Table~\ref{tab:input_ablation_flip_rates_by_feature} further support this observed pattern. Single-patch blurring rarely changes the final predicted class, but expanded target blurring produces substantially more prediction changes than expanded random blurring. Averaged across the eight selected features, expanded target interventions flip predictions in $22.0\%$ of cases, compared with $4.5\%$ for expanded random interventions. These results indicate that the image regions identified by selected SAE features are often prediction-relevant, even though the features themselves are not perfectly monosemantic.

\begin{table}[H]
\centering
\small
\begin{tabular}{rrrrr}
\toprule
Feature ID & Target & Random & Expanded Target & Expanded Random \\
\midrule
$1420$ & 0.02 & 0.00 & 0.54 & 0.06 \\
$1589$ & 0.04 & 0.02 & 0.38 & 0.10 \\
$2545$ & 0.00 & 0.00 & 0.04 & 0.02 \\
$2817$ & 0.00 & 0.00 & 0.24 & 0.02 \\
$3807$ & 0.00 & 0.02 & 0.16 & 0.06 \\
$3822$ & 0.00 & 0.00 & 0.06 & 0.02 \\
$3892$ & 0.02 & 0.00 & 0.22 & 0.02 \\
$6392$ & 0.04 & 0.00 & 0.12 & 0.06 \\
\bottomrule
\end{tabular}
\caption{Per-feature prediction flip rates for input-space ablations. Values denote the fraction of top-50 examples for each SAE feature whose predicted class changes after the corresponding blur intervention.}
\label{tab:input_ablation_flip_rates_by_feature}
\end{table}

\subsection{Feature-Space Ablation Results}

Feature-space ablations produce much smaller effects than input-space ablations. In the local setting, suppressing the selected SAE feature at only its top-activating spatial location has almost no effect on classifier outputs. The average true-class logit and classification margin drops are both only $0.007$. This suggests that locally suppressing a single SAE feature is not sufficient to substantially change the classifier's prediction.

Global feature suppression has a larger effect than local suppression, but the changes remain small compared with input-space blurring. Averaged across the eight selected features, global target suppression produces a true-class logit drop of $0.098$ and a classification margin drop of $0.113$, while global random suppression produces values close to zero. The per-feature global results show that feature ID $3892$ has a larger effect than others, but no selected feature produces prediction flips under feature-space suppression.\\

Feature-space drops are reported relative to the corresponding SAE reconstruction baseline. This is to separate changes caused by suppressing a specific SAE feature from distortions introduced by the SAE reconstruction process itself.

\begin{table}[H]
\centering
\small
\begin{tabular}{lrrrr}
\toprule
Metric & Local Target & Local Random & Global Target & Global Random \\
\midrule
True-class logit drop & 0.007 & 0.000 & 0.098 & 0.000 \\
True-class probability drop & \textless{}0.001 & 0.000 & \textless{}0.001 & \textless{}0.001 \\
Classification margin drop & 0.007 & 0.000 & 0.113 & 0.001 \\
Original top-1 confidence drop & \textless{}0.001 & 0.000 & \textless{}0.001 & \textless{}0.001 \\
Prediction flip rate & 0.000 & 0.000 & 0.000 & 0.000 \\
\bottomrule
\end{tabular}
\caption{A summary of feature-space ablation results averaged over the eight selected SAE features. Local interventions suppress the SAE feature only at its top-activating spatial location, while global interventions suppress the SAE feature across all spatial locations. Values with magnitude below $0.001$ are reported as $<0.001$.}
\label{tab:feature_ablation_summary}
\end{table}

\begin{table}[H]
\centering
\small

\begin{subtable}{\textwidth}
\centering
\small
\resizebox{\textwidth}{!}{
\begin{tabular}{rrrrrrrrr}
\toprule
Feature ID &
\multicolumn{2}{c}{True-class logit drop} &
\multicolumn{2}{c}{True-class prob. drop} &
\multicolumn{2}{c}{Classification margin drop} &
\multicolumn{2}{c}{Original top-1 conf. drop} \\
\cmidrule(lr){2-3}
\cmidrule(lr){4-5}
\cmidrule(lr){6-7}
\cmidrule(lr){8-9}
& Target & Random & Target & Random & Target & Random & Target & Random \\
\midrule
$1420$ & 0.001 & 0.000 & - \textless{}0.001 & 0.000 & -0.001 & 0.000 & \textless{}0.001 & \textless{}0.001 \\
$1589$ & 0.005 & 0.000 & 0.000 & - \textless{}0.001 & 0.001 & 0.000 & \textless{}0.001 & - \textless{}0.001 \\
$2545$ & 0.006 & 0.000 & 0.000 & 0.000 & 0.011 & 0.000 & \textless{}0.001 & 0.000 \\
$2817$ & 0.008 & 0.000 & 0.000 & 0.000 & 0.009 & 0.000 & \textless{}0.001 & 0.000 \\
$3807$ & 0.003 & 0.000 & - \textless{}0.001 & - \textless{}0.001 & 0.004 & 0.000 & - \textless{}0.001 & - \textless{}0.001 \\
$3822$ & 0.006 & 0.000 & 0.000 & 0.000 & 0.010 & 0.000 & \textless{}0.001 & 0.000 \\
$3892$ & 0.010 & 0.000 & 0.000 & 0.000 & 0.017 & 0.000 & \textless{}0.001 & 0.000 \\
$6392$ & 0.014 & 0.000 & - \textless{}0.001 & \textless{}0.001 & 0.002 & 0.000 & - \textless{}0.001 & - \textless{}0.001 \\
\bottomrule
\end{tabular}
}
\caption{Local SAE feature suppression. Target interventions suppress the selected SAE feature only at the top-activating spatial location, while random interventions suppress a randomly selected SAE feature at the same location.}
\label{tab:feature_ablation_local_by_feature}
\end{subtable}

\vspace{1.10em}

\begin{subtable}{\textwidth}
\centering
\small
\resizebox{\textwidth}{!}{
\begin{tabular}{rrrrrrrrr}
\toprule
Feature ID &
\multicolumn{2}{c}{True-class logit drop} &
\multicolumn{2}{c}{True-class prob. drop} &
\multicolumn{2}{c}{Classification margin drop} &
\multicolumn{2}{c}{Original top-1 conf. drop} \\
\cmidrule(lr){2-3}
\cmidrule(lr){4-5}
\cmidrule(lr){6-7}
\cmidrule(lr){8-9}
& Target & Random & Target & Random & Target & Random & Target & Random \\
\midrule
$1420$ & 0.012 & 0.001 & \textless{}0.001 & - \textless{}0.001 & 0.000 & 0.003 & 0.001 & \textless{}0.001 \\
$1589$ & 0.081 & 0.001 & 0.001 & - \textless{}0.001 & 0.033 & 0.003 & 0.001 & - \textless{}0.001 \\
$2545$ & 0.073 & 0.001 & \textless{}0.001 & 0.000 & 0.114 & 0.001 & \textless{}0.001 & 0.000 \\
$2817$ & 0.151 & -0.001 & \textless{}0.001 & - \textless{}0.001 & 0.160 & -0.003 & \textless{}0.001 & - \textless{}0.001 \\
$3807$ & 0.038 & 0.001 & - \textless{}0.001 & - \textless{}0.001 & 0.043 & -0.001 & - \textless{}0.001 & - \textless{}0.001 \\
$3822$ & 0.121 & 0.000 & \textless{}0.001 & \textless{}0.001 & 0.186 & 0.002 & \textless{}0.001 & \textless{}0.001 \\
$3892$ & 0.237 & 0.000 & 0.003 & \textless{}0.001 & 0.352 & -0.002 & 0.003 & \textless{}0.001 \\
$6392$ & 0.074 & 0.001 & - \textless{}0.001 & \textless{}0.001 & 0.016 & 0.001 & - \textless{}0.001 & \textless{}0.001 \\
\bottomrule
\end{tabular}
}
\caption{Global SAE feature suppression. Target interventions suppress the selected SAE feature across all spatial locations, while random interventions suppress a randomly selected SAE feature across all spatial locations.}
\label{tab:feature_ablation_global_by_feature}
\end{subtable}

\caption{Per-feature feature-space ablation results.}
\label{tab:feature_ablation_by_feature}
\end{table}

\begin{table}[H]
\centering
\small
\begin{tabular}{rrrrr}
\toprule
Feature ID & Local Target & Local Random & Global Target & Global Random \\
\midrule
$1420$ & 0.00 & 0.00 & 0.00 & 0.00 \\
$1589$ & 0.00 & 0.00 & 0.00 & 0.00 \\
$2545$ & 0.00 & 0.00 & 0.00 & 0.00 \\
$2817$ & 0.00 & 0.00 & 0.00 & 0.00 \\
$3807$ & 0.00 & 0.00 & 0.00 & 0.00 \\
$3822$ & 0.00 & 0.00 & 0.00 & 0.00 \\
$3892$ & 0.00 & 0.00 & 0.00 & 0.00 \\
$6392$ & 0.00 & 0.00 & 0.00 & 0.00 \\
\bottomrule
\end{tabular}
\caption{Per-feature prediction flip rates for feature-space ablations. No feature-space intervention changes the predicted class. Values denote the fraction of top-50 examples for each SAE feature whose predicted class changes after the corresponding feature-suppression intervention.}
\label{tab:feature_ablation_flip_rates_by_feature}
\end{table}

Overall, the feature-space ablation results suggest that the selected SAE features are not individually decisive causal units in the classifier representation. After accounting for SAE reconstruction, suppressing a single sparse feature can slightly reduce classifier evidence, especially when the feature is suppressed across all spatial locations, but it does not substantially alter the final decision. Together with the input-space results, this suggests that SAE features identify prediction-relevant image regions, while the classifier's internal evidence remains distributed across multiple features and spatial locations.

	\section{Discussion and Limitations}

The results suggest that sparse autoencoders can expose partially interpretable features in the late representations of a fine-tuned ConvNeXt aircraft classifier. Qualitative inspection shows that several SAE features have recognizable and dominant visual themes. These features are not perfectly monosemantic, but they often correspond to aircraft-related cues that are plausible evidence for aircraft image classification.\\

The input-space ablation results provide evidence that the regions associated with selected SAE features are prediction-relevant. Blurring SAE-selected patches generally produces larger reductions in classifier evidence than blurring randomly selected patches from the same image. This effect is especially clear in the expanded-region setting, where the $3 \times 3$ neighborhood around the selected patch is ablated. The stronger effect of expanded-region ablation suggests that the visual evidence captured by an SAE feature is often not confined to a single spatial location, but instead involves the surrounding image context.\\

In contrast, feature-space ablations produce much smaller effects. Suppressing a selected SAE feature at only its top-activating spatial location has almost no effect on classifier outputs, and even suppressing the feature globally across all spatial locations produces only modest changes in logits and margins. None of the selected feature-space interventions changes the final prediction. This contrast between input-space and feature-space ablations suggests that the SAE features are useful for identifying prediction-relevant image regions, but individual sparse features are not necessarily standalone causal units in the classifier representation in this experimental setup.

A likely explanation for this is that aircraft recognition depends on distributed evidence, where aircraft distinctions can involve multiple visual cues. A single SAE feature may capture one component of this evidence, but the classifier may rely on several features and spatial regions in conjunction. This is also consistent with the observed polysemanticity of the selected features: most features have a dominant visual theme, but also activate on related nearby structures or broader part-context combinations.\\

While these results are informative, several limitations in this study should be noted. First, the analysis is restricted to one classifier model architecture, one dataset, and one SAE trained on a single activation layer. The results may differ for other vision backbones, earlier or later layers, larger SAEs, or different datasets. Second, the qualitative interpretations are based on manual inspection of top-activating patches. Although this is standard in interpretability work, the assigned labels are approximate and may miss subtler patterns in the feature activations. Third, the ablation study uses a small manually selected set of just eight SAE features, rather than evaluating all learned features systematically. These features were chosen to cover a range of aircraft-related cues, but they may not be representative of the full SAE dictionary.

There are also limitations in the ablations themselves. The blurring performed for input-space ablation is an approximate intervention: it removes local visual detail, but may also introduce distribution shift or affect nearby visual evidence. Expanded-region blurring produces stronger effects, but it is less spatially precise because it removes a larger part of the image. Feature-space ablation is also limited because it suppresses individual SAE features independently. If classifier evidence is distributed across multiple sparse features, then zeroing one feature may underestimate the causal role of the broader concept.\\

Future work could extend this analysis by ablating groups of related SAE features, evaluating larger sets of features automatically, comparing multiple model architectures, and applying the method to other tasks.

	\section{Conclusion}

This work investigates whether sparse autoencoders can be used to interpret intermediate representations of a vision model trained for aircraft classification. We train a ConvNeXt classifier on FGVC-Aircraft, extract patch-level activations from its final feature stage, and train an SAE on these activations. The learned sparse features reveal partially interpretable aircraft-related patterns. Ablation experiments show that the image regions associated with selected SAE features are often prediction-relevant. Blurring SAE-selected regions reduces classifier evidence more than blurring random regions, especially when the surrounding $3 \times 3$ neighborhood is ablated. However, direct feature-space suppression produces much smaller effects and does not change the classifier's final predictions. These results suggest that SAE features can help identify meaningful visual evidence used by the classifier, while also showing that individual sparse features are not necessarily clean and independently decisive causal units.\\

Overall, the findings support the use of sparse autoencoders as a tool for probing representations in vision models. At the same time, they highlight important challenges: learned SAE features may be polysemantic, context-dependent, and only partially aligned with human-defined object parts. This suggests that future interpretability work on vision models should analyze not only whether sparse features are visually interpretable, but also how they interact with distributed classifier evidence across features and spatial locations.

	\clearpage
	\bibliographystyle{unsrtnat}
	\bibliography{main}
	
\end{document}